\documentclass[letterpaper, 10 pt, conference]{ieeeconf}  

\IEEEoverridecommandlockouts                              

\title{\LARGE \bf
STAR: A Foundation Model-driven Framework for Robust Task Planning and Failure Recovery in Robotic Systems
}

\author{Md Sadman Sakib and Yu Sun
\thanks{The authors are affiliated with the Robot Perception and Action Lab (RPAL) under the Department of Computer Science \& Engineering at the University of South Florida, Tampa, FL, USA.
        {\tt\small mdsadman@usf.edu, yusun@usf.edu}}%
}

\usepackage{graphicx}
\usepackage{soul}

\usepackage{hyperref}
\let\labelindent\relax
\usepackage[inline]{enumitem}
\usepackage{cite}

\begin{document}

\maketitle
\thispagestyle{empty}
\pagestyle{empty}

\begin{abstract}

Modern robotic systems, deployed across domains from industrial automation to domestic assistance, face a critical challenge: executing tasks with precision and adaptability in dynamic, unpredictable environments. To address this, we propose STAR (Smart Task Adaptation and Recovery), a novel framework that synergizes Foundation Models (FMs) with dynamically expanding Knowledge Graphs (KGs) to enable resilient task planning and autonomous failure recovery. 
While FMs offer remarkable generalization and contextual reasoning, their limitations—including computational inefficiency, hallucinations, and output inconsistencies—hinder reliable deployment. STAR mitigates these issues by embedding learned knowledge into structured, reusable KGs, which streamline information retrieval, reduce redundant FM computations, and provide precise, scenario-specific insights. The framework leverages FM-driven reasoning to diagnose failures, generate context-aware recovery strategies, and execute corrective actions without human intervention or system restarts. Unlike conventional approaches that rely on rigid protocols, STAR dynamically expands its KG with experiential knowledge, ensuring continuous adaptation to novel scenarios.
To evaluate the effectiveness of this approach, we developed a comprehensive dataset that includes various robotic tasks and failure scenarios. Through extensive experimentation, STAR demonstrated an 86\% task planning accuracy and 78\% recovery success rate, showing significant improvements over baseline methods. The framework's ability to continuously learn from experience while maintaining structured knowledge representation makes it particularly suitable for long-term deployment in real-world applications.

\end{abstract}

\section{Introduction}

Robotic systems are increasingly deployed in dynamic, unstructured environments—from precision-driven industrial automation to human-centric domestic assistance and extreme scenarios like space exploration \cite{Asgharian2022ARO, paulius2019survey}. While these domains demand autonomous adaptability, traditional robotic architectures, reliant on rigid rule-based programming, often fail to address the unpredictability of real-world tasks. Critical challenges persist in enabling robots to autonomously recover from execution failures (e.g., sensor errors, grasping inaccuracies, or environmental changes) without human intervention or costly task restarts \cite{Huang2024ReKepSR, liu2023reflect, Skreta2024RePLanRR}. This gap underscores the need for systems that integrate high-level reasoning with structured, reusable knowledge to achieve both flexibility and reliability.

Recent advances in foundation models (FMs) offer promising avenues for robotic task planning due to their ability to parse natural language instructions and generalize across tasks. However, FM-centric approaches face inherent limitations: they are prone to hallucination, lack domain-specific precision, and suffer from computational inefficiencies when repeatedly processing similar scenarios. These shortcomings become acute in failure recovery, where context-aware diagnosis and precise corrective actions are critical. For instance, while FMs might suggest plausible recovery strategies, they often lack the structured reasoning to prioritize solutions based on historical success rates or domain constraints.

\begin{figure}[ht!]
	\centering
	\includegraphics[width=\columnwidth]{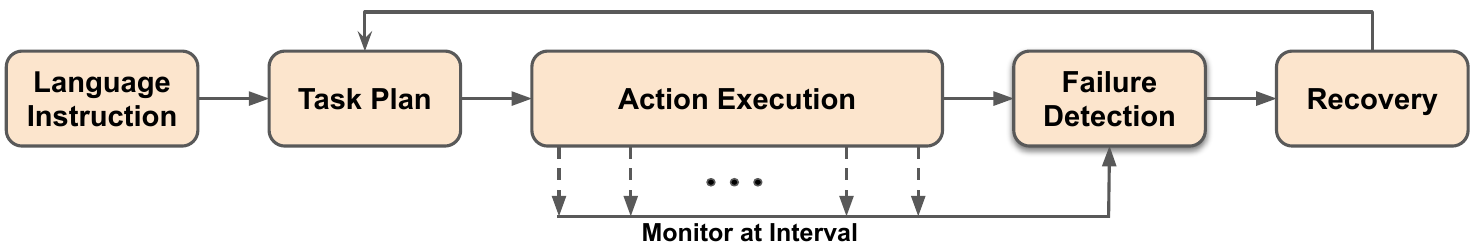}
	\caption{Overview of our task planning and failure recovery system.}
	\label{fig:c5-intro}
\end{figure}

To address these challenges, we propose STAR (Smart Task Adaptation and Recovery), a framework that synergizes the contextual reasoning of FMs with a dynamically expanding knowledge graph (KG) to enable robust task execution and autonomous failure recovery. As illustrated in Figure \ref{fig:c5-intro}, STAR transforms natural language commands into executable plans while continuously monitoring execution. When failures occur, the system diagnoses root causes by correlating real-time sensor data with KG-stored knowledge—such as historical failure patterns, object affordances, and environmental constraints—to generate context-optimized recovery strategies. Crucially, the KG evolves with each interaction, capturing new failure modes and solutions to enhance future adaptability.

This hybrid design mitigates FM limitations by offloading structured knowledge storage and retrieval to the KG, reducing redundant computations and grounding FM outputs in domain-specific facts. For example, while an FM might propose multiple recovery actions, the KG prioritizes those validated in similar past scenarios, ensuring efficiency and reliability. Unlike traditional systems that rely on predefined recovery protocols, STAR dynamically adapts strategies to novel failures, such as substituting unavailable tools or adjusting motion trajectories to avoid newly detected obstacles.

Comprehensive experiments, conducted on a dataset encompassing diverse robotic tasks and failure scenarios, demonstrate the effectiveness of this approach. The results show significant improvements in task completion and the ability to recover from unforeseen challenges. Our contributions and outcomes in this paper are as follows:

\begin{enumerate*}[label=(\roman*)]
\item A hybrid FM-KG architecture that balances generalizable reasoning with structured knowledge retention for robust task planning;
\item An adaptive recovery mechanism that autonomously diagnoses failures, selects context-aware solutions from the KG, and updates the knowledge base with new insights;
\item A publicly shared dataset of robotic tasks, failure scenarios, and recovery benchmarks to standardize future research;
\item Empirical validation showing STAR’s superiority in handling novel failures, with quantifiable gains in resilience and efficiency.
\end{enumerate*}

\section{Background and Related Works}

\subsection{Knowledge Graph}
A Knowledge Graph (KG) is a structured framework that organizes entities and their interrelationships in a graph format, facilitating context-aware reasoning and efficient retrieval of procedural knowledge. In this work, we use the Functional Object-Oriented Network (FOON) \cite{paulius2016functional} as our KG, which models manipulation action sequences using object and motion nodes. The core unit of FOON is the functional unit, representing a single action that transforms input object nodes into output object nodes. For example, a cutting action might transform a whole vegetable into sliced pieces. FOON currently includes 140 recipes, represented as a graph of functional units. We represent high-level task plans as task trees, which are sequences of functional units that, when executed in order, lead the robot from an initial state to the goal.  For more details on task tree representation, we refer the reader to \cite{Sakib2022ApproximateTT, consolidating}.


\subsection{Classical Planning}
Classical planning forms the foundation of artificial intelligence for generating action sequences to transition from an initial state to a goal state \cite{McCarthy1963SituationsAA}. It assumes a fully observable and deterministic world, enabling the use of algorithms that produce optimal or near-optimal solutions. Early implementations, like Shakey the robot \cite{Nilsson1984ShakeyTR}, leveraged STRIPS \cite{Fikes1971STRIPSAN} to plan in structured environments, laying the groundwork for modern robotics applications.

Subsequent advancements, such as PRODIGY \cite{Carbonell1991PRODIGYAI} and Hierarchical Task Networks (HTNs) \cite{Nau}, extended the versatility of classical planning, while languages like PDDL \cite{mcdermott1998pddl, Fox2003PDDL21AE} and ASP \cite{Lifschitz2002AnswerSP} provided structured frameworks for defining complex planning domains. Algorithms like A* \cite{Hart1968AFB}, GraphPlan \cite{Chen2021GraphPlanSG}, and Fast Forward (FF) \cite{hoffmann2001ff} introduced heuristic strategies to improve planning efficiency.
More recent approaches, like Task and Motion Planning (TAMP) \cite{Kaelbling2013IntegratedTA, Lagriffoul2018PlatformIndependentBF}, combine discrete task planning with continuous motion planning to address real-world complexities.While classical planning is robust for structured environments, its reliance on deterministic assumptions limits its adaptability to dynamic and uncertain settings. 

\subsection{Language Model-Based Task Planning}
The advent of large language models (LLMs) such as GPT-3 \cite{Brown2020LanguageMA} and GPT-4 \cite{Achiam2023GPT4TR} has transformed task planning. Leveraging the Transformer architecture \cite{Vaswani2017AttentionIA}, LLMs exhibit exceptional zero-shot and few-shot generalization, drawing on vast pretraining datasets to generate and adapt complex plans for long-horizon robotic tasks. For example, Erra \cite{erra} utilized LLMs for robotic manipulation, while Huang et al. \cite{Huang2022LanguageMA} demonstrated their ability to dynamically adapt plans. Shah et al. \cite{Shah2022LMNavRN} showcased LLM-driven reasoning in robotic navigation within unstructured environments.

Despite their strengths, LLMs require grounding to accommodate physical constraints and robot-specific capabilities. Frameworks like SayCan \cite{saycan} align LLM plans with a robot's skills and context, while Text2Motion \cite{lin2023text2motion} and ProgPrompt \cite{singh:progprompt} integrate LLMs with policy networks for real-world execution. Tools like Reflect \cite{liu2023reflect} enhance planning by incorporating multimodal inputs, translating sensor data into textual prompts. LLM+P \cite{LLMP} combines LLM-generated PDDL definitions with classical planners for structured solutions, and ToolFormer \cite{Schick2023ToolformerLM} augments LLMs with APIs for real-time data retrieval, enhancing planning efficiency and adaptability.

\subsection{Multimodal Foundation Models}
Advancements in LLMs have driven the development of vision-language models (VLMs) \cite{Luo2024VisionLanguageMF, Liang2023CrowdCLIPUC}, which integrate visual and textual data through crossmodal connectors \cite{Li2021GroundedLP}. This alignment enables VLMs to process and understand both images and text, making them effective for tasks like visual question answering \cite{He2023AnnoLLMML} and image captioning \cite{Kabra2023LeveragingVP}. Our research, however, leverages VLMs for complex long-horizon planning in robotics by exploiting their visually grounded knowledge and world understanding.

Foundational models such as CLIP \cite{Radford2021LearningTV} highlight the integration of vision and language using large-scale image-text pairs to establish semantic relationships. CLIP’s contrastive learning approach enables zero-shot capabilities, while enhanced models like BLIP \cite{Li2022BLIPBL}, CLIP2 \cite{Zeng2023CLIP2CL}, and FLIP \cite{Srivatsan2023FLIPCF} refine these capabilities through improved alignment, masking techniques, and efficient training.

While VLMs are often used for perception-focused applications, our study extends their utility to high-level planning in robotics. This approach moves beyond generating visual representations for control tasks by integrating broader reasoning and open-world capabilities, as seen in models like PaLM-E \cite{Driess2023PaLMEAE}. Unlike PaLM-E, our method avoids environment-specific retraining, maintaining zero-shot adaptability across diverse tasks. Practical applications of VLMs in robotics include
ViLaIn \cite{shirai2024visionlanguage}, which generates problem files from observations and reprompts LLMs when necessary. Additionally, interactive planning methods \cite{sun2023interactive} enhance adaptability by allowing LLMs to address partially observable environments.

\subsection{Adaptive Error Management}
Effective error handling not only enhances system resilience but also strengthens human-robot collaboration by making robots more reliable and dependable partners.
A significant challenge in this domain is executing complex, constraint-heavy tasks over extended periods. Faithful Long-Horizon Task Planning \cite{faithful} addresses this by decomposing tasks into smaller, manageable sub-tasks, leveraging long-term and short-term memory frameworks. 

The CAPE framework \cite{raman2023cape} refines task plans by addressing precondition errors identified in simulation. While effective in virtual settings, translating these corrections to real-world scenarios remains challenging without integrating multimodal sensory inputs. Similarly, AutoTAMP \cite{chen2024autotamp} focuses on identifying semantic and syntactic errors in task plans. By generating intermediate representations using Signal Temporal Logic (STL) and validating them with motion planners, AutoTAMP ensures alignment between robotic actions and initial instructions.
Tree-Planner \cite{hu2023treeplanner} introduces an optimization approach that consolidates multiple task plans into a single aggregated graph.

In Human-Robot Interaction (HRI), making robot failures interpretable to users is crucial for building trust and enabling non-experts to assist robots effectively. Research emphasizes the importance of failure explanations in fostering transparency and collaboration \cite{Ye2019HumanTA}. Existing works have explored specific failure scenarios, such as object-picking errors \cite{Das2021ExplainableAF}, pick-and-place failures \cite{Diehl2022WhyDI}, navigation issues \cite{Song2022LLMPlannerFG}, and short-horizon manipulation breakdowns \cite{Inceolu2020FINONetAD}. In contrast, our framework leverages the advanced reasoning capabilities of LLMs to detect and explain a diverse range of failure types without relying on task-specific assumptions.

\section{Task Planning Methodology}

This section introduces our hybrid task planning framework STAR, which integrates KGs and LLMs. STAR balances the strengths of both: KGs provide a reliable and reusable knowledge structure, while LLMs offer adaptability to novel scenarios. This synergy ensures robust and flexible task planning, enabling robots to handle diverse tasks effectively.

\begin{figure}[ht!]
  \centering
  \includegraphics[width=\columnwidth]{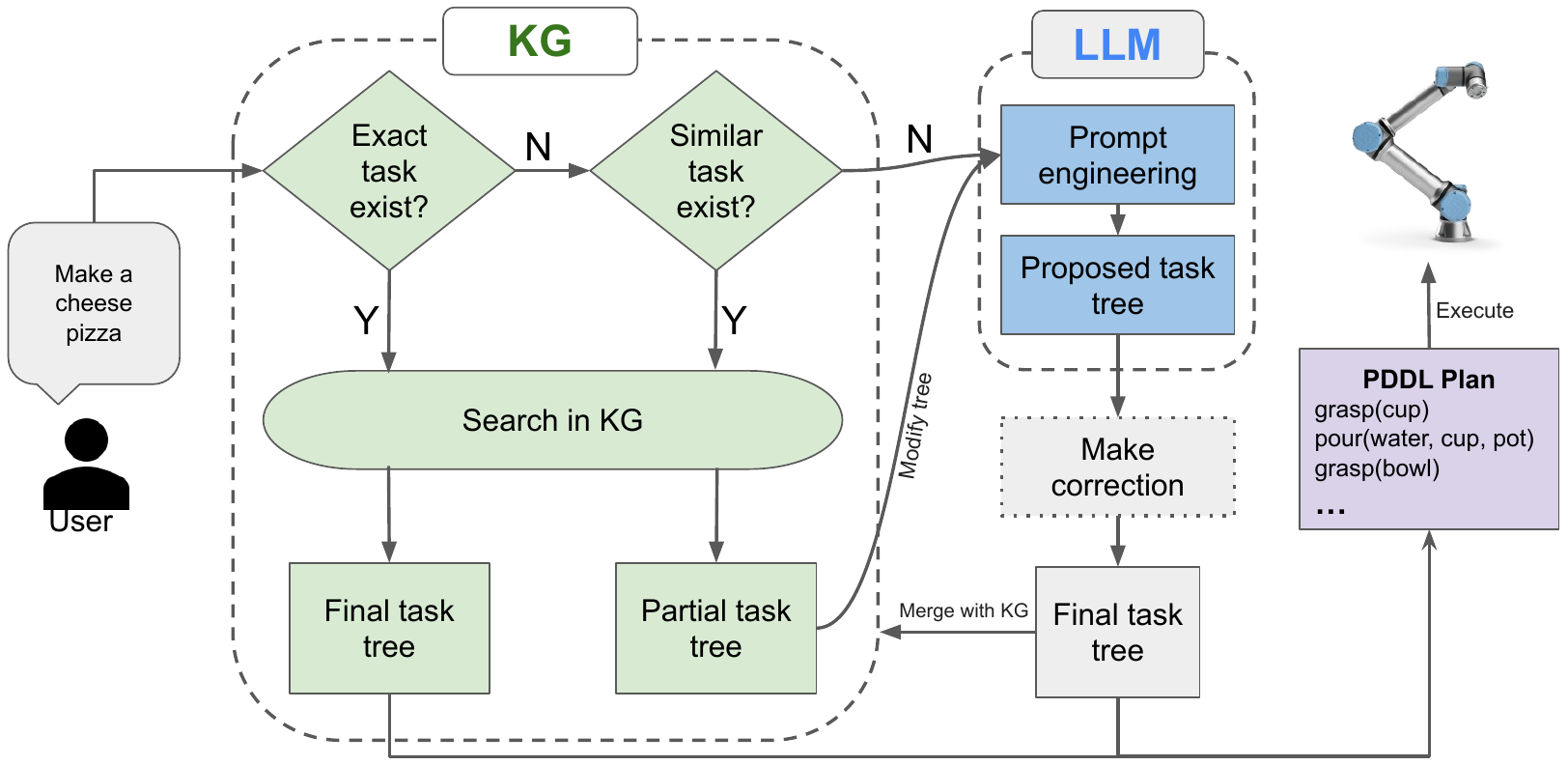}
  \caption[Overview of our task planning procedure.]{Task planning pipeline in STAR. Given a natural language input, the system searches for matching tasks in the Knowledge Graph (KG). If no match exists, LLM generates a new task tree, which is verified, converted to PDDL format, and stored in the KG for future use.}
  \label{fig:c5-tree-generation}
\end{figure}

The motivation arises from the challenges of depending solely on KGs or LLMs. While KGs provide structured and reliable knowledge, they are limited to specific domains and require significant effort to build. LLMs, in contrast, are versatile but prone to errors and inconsistencies. To overcome these issues, we adopt a hierarchical task planning approach inspired by human problem-solving. High-level planning outlines a general strategy, and low-level planning refines it into precise, actionable steps. This structure ensures the system remains flexible for new tasks and dependable for known ones, with lifelong learning gradually decreasing reliance on LLMs. An overview of the proposed method is outlined in Figure \ref{fig:c5-tree-generation}.

\subsection{Task Tree Retrieval from KG} 
Although KGs excel in structured data representation, they lack the ability to process natural language, making it challenging to directly interpret user instructions. To address this limitation, we integrate a Foundation Model (FM) to interpret user inputs and determine the goal node—often the name of a dish in culinary tasks—which serves as the target endpoint for KG-based search. Complementing this, the robot's vision system identifies objects and their states within the environment to aid in execution. Multiple scenarios may arise during the KG search:

\begin{itemize}
\item Case 1: KG lacks a corresponding task tree for the requested dish category. A recipe classification system assists in categorizing recipes into 30 dish classes, simplifying the process of identifying similar dishes within FOON.

\item Case 2: An exact match is unavailable, but recipes from the same category are present, such as having corn soup instead of lentil soup. In this case, the KG provides a close approximation.

\item Case 3: KG contains an exact match for the requested dish, offering a complete task tree.
\end{itemize}

\subsection{LLM-Assisted Task Tree Generation}

For Cases 1 and 2, we use LLMs to create or adjust task trees. If a partial task tree is available, we include it in the prompt with the user's request, allowing the LLM to modify it. If there is no partial tree, we use few-shot prompting to help the LLM generate a new task tree. Few-shot prompting is needed because the LLM is not familiar with the task tree structure. By giving it a few examples of FOON task trees, we teach it the syntax, not the semantic knowledge. Our experiments show that giving five examples is sufficient.

\subsection{Lifelong Learning and Knowledge Expansion}

Our architecture capitalizes on the repetitive nature of robotic tasks, such as cleaning, by promoting lifelong learning. Over time, successfully executed plans are stored for future reuse, reducing reliance on probabilistic LLMs and leveraging the accumulated structured knowledge in the KG. If modifications to a plan are necessary, corrections are visualized and merged with the existing KG, preventing duplicate entries and expanding the functional units repertoire. Consequently, this ongoing knowledge expansion decreases the need for LLM queries, enhancing efficiency, accuracy, and cost-effectiveness.

\subsection{Low-Level Plan Generation}

\begin{figure}[ht!]
  \centering
  \includegraphics[width=\columnwidth]{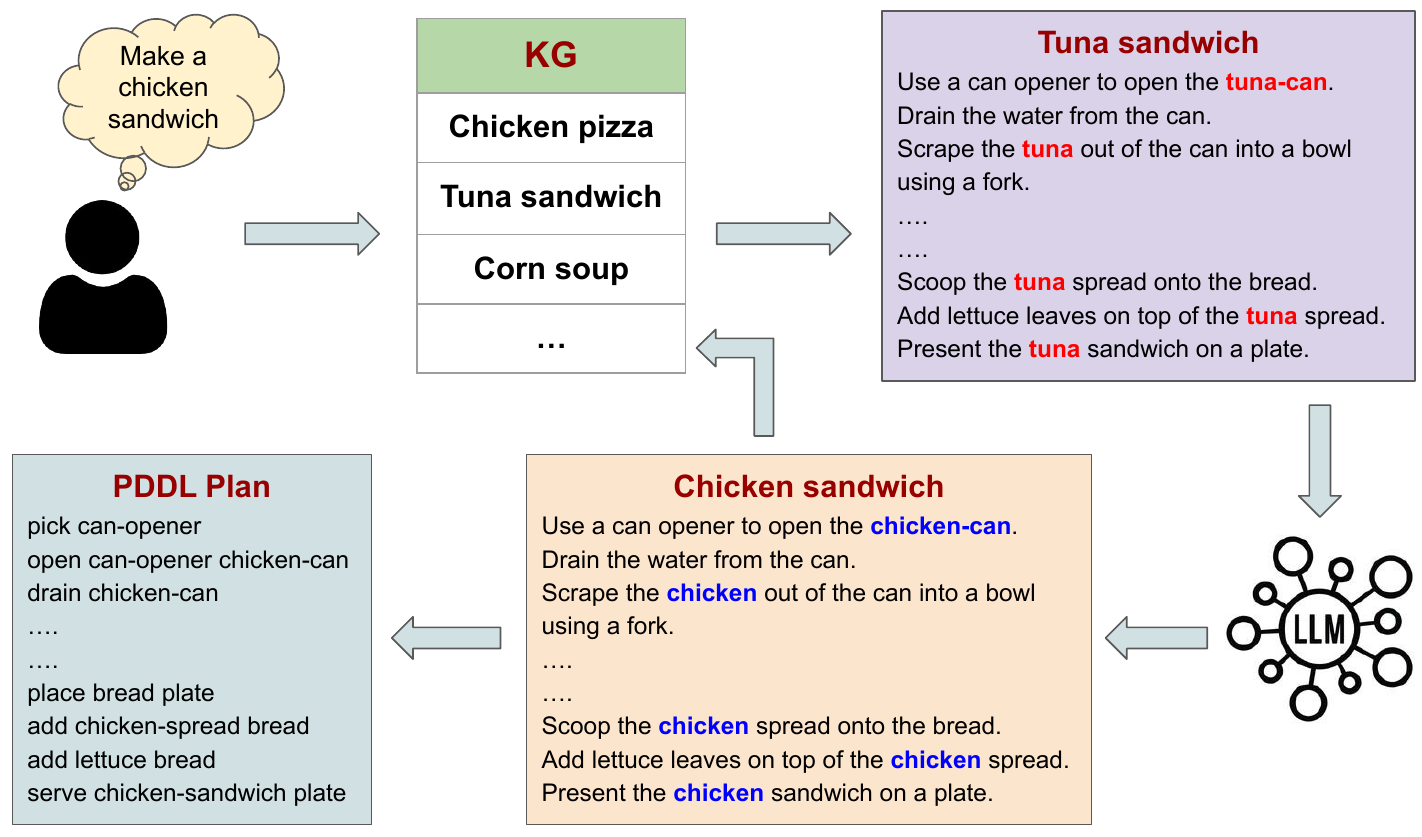}
  \caption{An example of task planning using STAR. For better readability, the functional units in the task tree are translated to natural language sentences.}
  \label{fig:c5-tree-example}
\end{figure}

The final step in our task planning process is converting the completed task tree—whether from the KG, LLM, or a combination of both—into a PDDL plan for robotic execution. We use simple prompts to generate a PDDL domain and problem file for each functional unit, guiding the conversion into PDDL format. Our experiments show that LLMs are highly accurate in this conversion. After that, a classical planner like Fast Forward is used to solve the planning problem, ensuring precise task execution. Figure \ref{fig:c5-tree-example} illustrates our task planning pipeline, showing how the task tree is generated from instructions and converted into a PDDL plan.
 
\section{Failure Recovery}

Robotic systems in unstructured environments are prone to execution failures due to unpredictable real-world conditions, including planning errors, execution mistakes, and perception issues. STAR focuses on execution errors during task performance, aiming to detect failures early and implement adaptive recovery strategies to ensure task completion. While existing literature addresses specific failure types—such as object-picking \cite{Das2021ExplainableAF, Das2021SemanticBasedEA}, pick-and-place tasks \cite{Diehl2022WhyDI}, and navigation failures \cite{Song2022LLMPlannerFG, 10403995}—most frameworks are task-specific. In contrast, STAR uses Large Language Models (LLMs) for broad failure detection and explanation, without relying on task-specific assumptions.

STAR emphasizes continuous monitoring during task execution to detect unsafe actions or unintended events early, reducing critical failures. This monitoring is crucial for two reasons:

\begin{itemize} 
\item \textit{Safety}: Preventing unsafe actions, like turning on a stove, which may cause harm. 
\item \textit{Collateral Events}: Avoiding unintended interactions with objects, like knocking over a glass while preparing a dish. Traditional planners focus only on task-relevant objects, often missing these environmental impacts. Our approach incorporates interval-based monitoring to ensure safety and enable quick recovery from unexpected events. 
\end{itemize}

\subsection{Failure Detection through Vision-Language Models (VLM)}
\begin{figure}[ht!]
	\centering
	\includegraphics[width=\columnwidth]{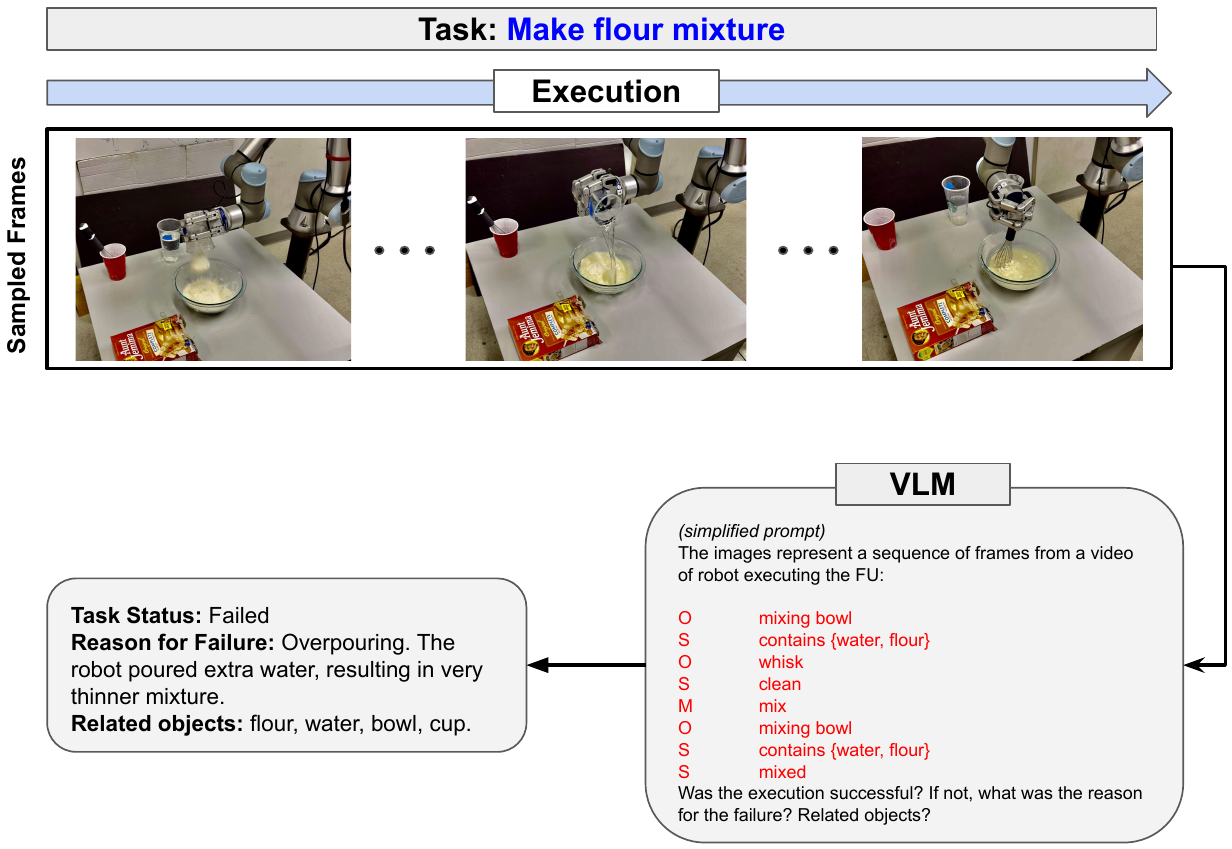}
	\caption{Failure detection pipeline.}
	\label{fig:c5-detection}
\end{figure}

STAR employs GPT-4 Vision, to detect failures by analyzing frames sampled periodically during the robot’s execution. Instead of waiting until task completion, we capture intermediate states to identify both subtle and major issues. 
The number of sampled frames varies based on the task’s duration. For actions lasting less than five minutes—common in tasks like object manipulation or cooking—ten frames are usually sufficient for reliable detection.
Existing approaches that rely on final-frame analysis may fail to detect the root cause of a failure or ignore events occurring mid-execution. By providing multiple frames along with contextual information about the functional unit being executed, the VLM can detect failures more precisely. 
Figure~\ref{fig:c5-detection} illustrates our failure detection pipeline. For instance, while preparing a pancake mixture, the robot overpours water into the bowl, making the mixture too watery. The VLM detects the overpouring error mid-execution, allowing corrective actions to be taken immediately.

\subsection{Recovery Mechanism}
When a failure is detected, the system determines whether it can be resolved through re-execution or if the task plan needs modification.

\begin{itemize}
    \item \textit{Re-execution without Plan Modification} - some failures can be addressed by simply repeating the action. For example, if the robot fails to grasp a bottle and the bottle remains in place, the same grasp action can be retried.
    \item \textit{Plan Adjustment and Adaptive Recovery} - in more complex scenarios, a change in the task plan is required. For instance, in the overpouring example mentioned earlier, adding more flour to balance the mixture would be necessary. These adjustments ensure the task can still be completed successfully without starting over.
\end{itemize}

\subsection{Hierarchical Re-planning with FailNet}
Our failure recovery system adopts a hierarchical re-planning approach, leveraging both high-level and low-level planning to adapt to unforeseen events. The recovery plan generation process mirrors the task planning approach described earlier but with a specialized focus on handling failures.

\begin{itemize}
    \item \textit{High-Level Recovery Plan} - 
we search a knowledge graph called FailNet, which has a similar structure to the FOON. However, FailNet specifically stores functional units (FUs) related to failure recovery strategies. For example, it contains subgraphs indicating how to recover from an overpouring event by adding extra ingredients or how to correct grasp failures.

FailNet is initialized with a minimal set of failure recovery strategies and grows over time as new recovery actions are generated by LLMs. After a successful recovery, the LLM-generated plan is added back to FailNet to improve future response efficiency. This continuous learning ensures that the robot becomes increasingly proficient at handling failures.

\item \textit{Low-Level Recovery Plan} - 
once a high-level recovery plan is generated, it is converted into a PDDL plan for robotic execution. This PDDL plan ensures that the recovery actions are executed precisely and efficiently.
Figure~\ref{fig:c5-replan} presents an overview of the recovery process. Whenever a failure is detected, the system initiates the recovery pipeline, beginning with a high-level search in FailNet and, if necessary, generating new recovery strategies through LLM assistance.

\begin{figure}[ht!]
	\centering
	\includegraphics[width=\columnwidth]{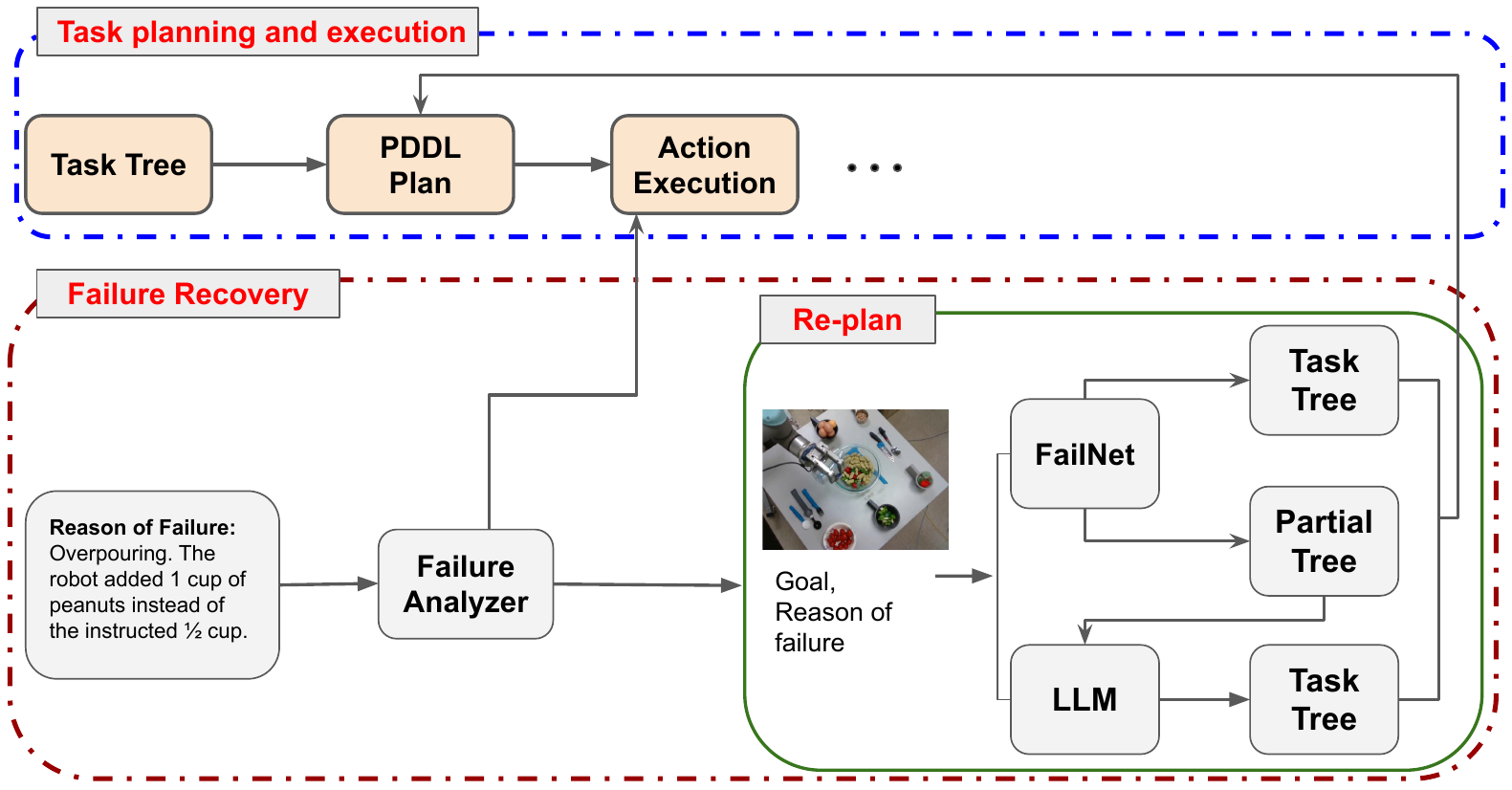}
	\caption{Failure recovery method.}
	\label{fig:c5-replan}
\end{figure}
\end{itemize}

\subsection{FailNet Knowledge Expansion and Lifelong Learning}
\begin{figure}[ht!]
	\centering
	\includegraphics[width=\columnwidth]{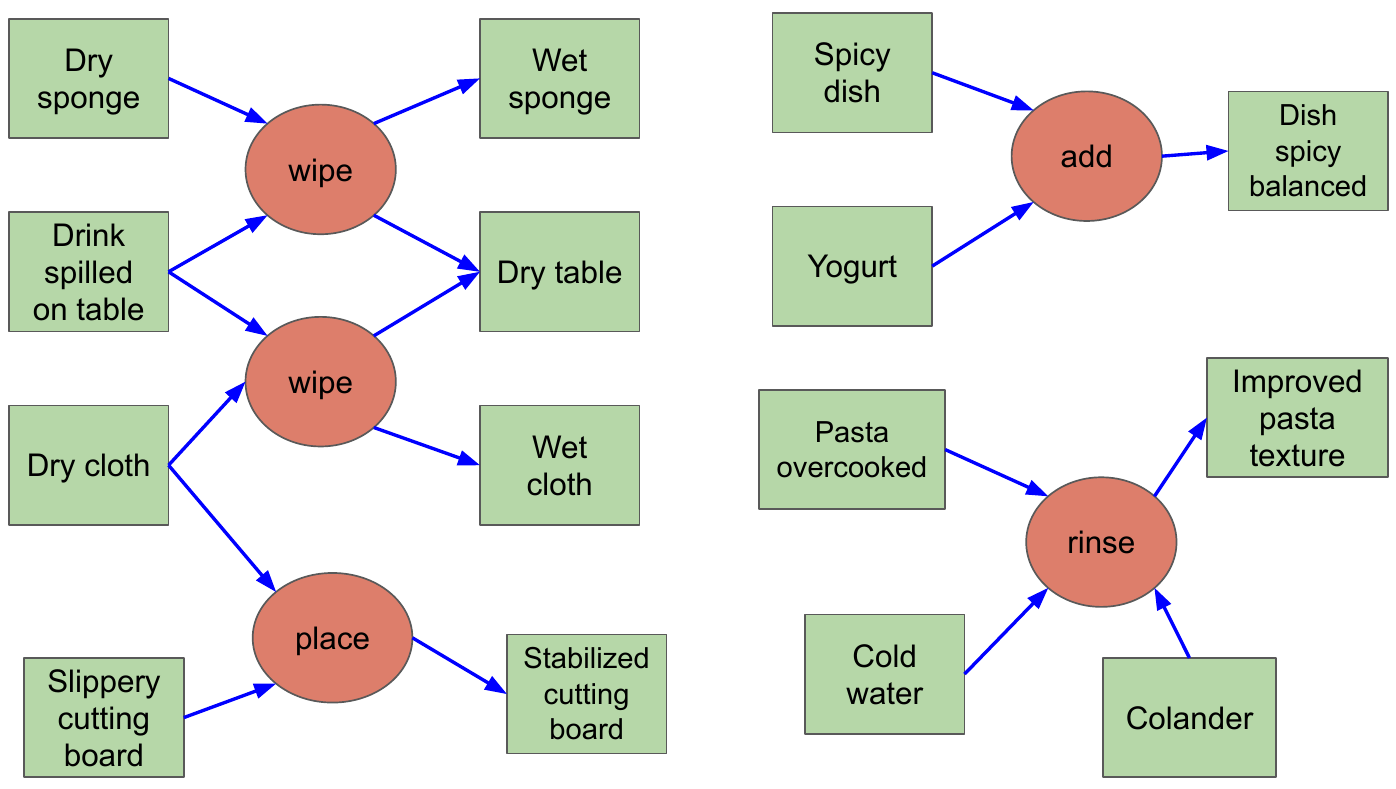}
	\caption{Snapshot from FailNet}
	\label{fig:c5-failnet}
\end{figure}

A key feature of our framework is lifelong learning. FailNet starts with limited recovery knowledge but grows continuously as the robot encounters and resolves new failures. When an LLM generates a new recovery plan that succeeds, the corresponding functional units are added to FailNet. This ensures that similar failures can be handled more efficiently in the future without repeatedly querying the LLM.
Figure~\ref{fig:c5-failnet} shows a snapshot of FailNet, highlighting how new recovery strategies are incrementally integrated. FailNet's structure is designed to merge seamlessly with FOON, but it is kept separate to reduce search complexity.

\section{Experiments and Results}

We evaluate the effectiveness of STAR in two primary areas: task planning and failure recovery. Each component is assessed separately to measure performance and ensure robust execution under real-world conditions. 

\subsection{Task Planning Evaluation}
\begin{figure}[ht!]
	\centering
	\includegraphics[width=\columnwidth]{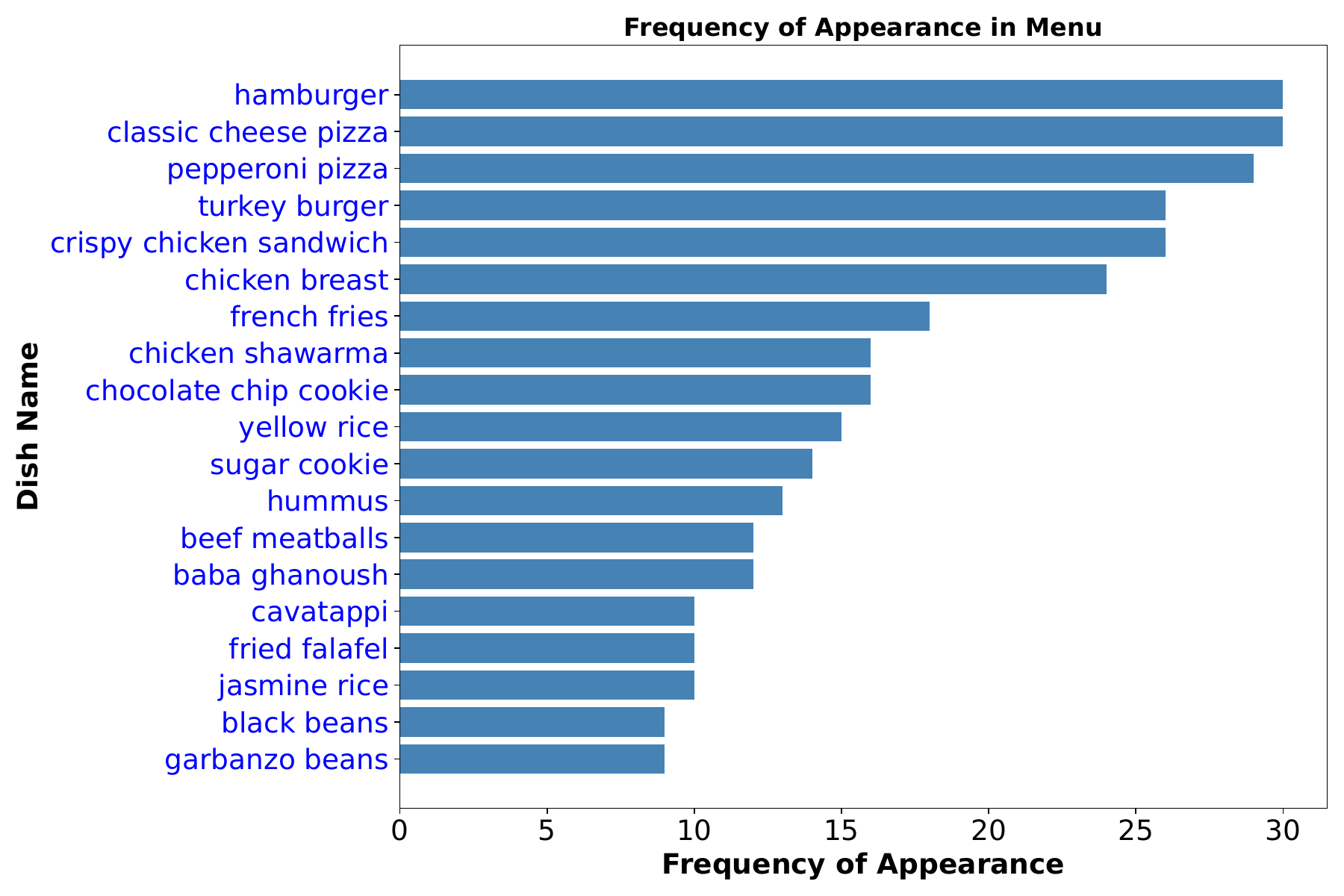}
	\caption{Frequency of dishes featured on the cafeteria menu}
	\label{fig:c5-cafe-menu}
        \vspace{-0.15in}
\end{figure}

Given that robots often perform repetitive tasks daily, the university cafeteria provides an ideal environment to test the robot’s task-planning capabilities. We curated a dataset containing the daily menu over the span of one month, ensuring a variety of dishes and scenarios. The goal is to determine how efficiently the robot can generate task plans for the dishes listed. Figure \ref{fig:c5-cafe-menu} shows a visualization of the most frequently occurring dishes in the cafeteria’s July 2024 menu.

For every dish in the dataset, the robot generates a task tree outlining the necessary steps. These task trees are evaluated using the progress line framework, which provides insights into how well the robot progresses through each task. To assess the model’s performance, we compare it with several baseline methods, including FOON-based approaches and GPT-4-driven systems. The results are summarized in Table \ref{tab:accuracy_comparison}.

\begin{table}[ht!]
\centering
\caption{Comparison of task planning accuracy across methods}
\label{tab:accuracy_comparison}
\begin{tabular}{|l|c|}
\hline
\textbf{Method}            & \textbf{Accuracy (\%)} \\ \hline
FOON \cite{paulius2016functional}                      & 5  \\ \hline
FOON + Substitution \cite{Sakib2022ApproximateTT}      & 54 \\ \hline
GPT-4 \cite{Achiam2023GPT4TR}                    & 78 \\ \hline
Tree Consolidation \cite{consolidating}        & 81 \\ \hline
\textbf{STAR}                & \textbf{86}  
\\ \hline
\end{tabular}
\end{table}

A key benefit of our approach is the ability to store corrected plans. Unlike other methods, which may repeat past mistakes even after correction, our system guarantees that, once a plan is corrected, it will produce the correct result every time it encounters the same query. This feature is especially advantageous when dealing with repetitive tasks across the month.

\subsection{Failure Detection and Recovery}
To test the framework's failure detection and recovery capabilities, we created a comprehensive video dataset consisting of robotic task failures. These videos were compiled from YouTube demonstrations, robotics competitions, and in-lab experiments. The dataset focuses primarily on pouring, mixing, and pick-and-place tasks, which frequently lead to execution failures. Table \ref{tab:video_distribution} provides an overview of the video sources and task types included in the dataset.

\begin{table}[ht]
    \centering
    \caption{Distribution of video sources and types in the dataset.}
    \begin{tabular}{|l|c|l|c|}
        \hline
        \textbf{Source of Videos} & \textbf{Count} & \textbf{Type of Videos} & \textbf{Count} \\ \hline
        YouTube                  & 23             & Pouring                 & 36             \\ 
        Competition              & 23             & Mixing                  & 19             \\ 
        Lab Experiment           & 55             & Pick and Place          & 23             \\ 
                                &                & Others                  & 23             \\ \hline
    \end{tabular}
    \label{tab:video_distribution}
    \vspace{-0.1in}
\end{table}

Some examples of failures in the dataset include:

\begin{itemize}
    \item \textit{Overpouring -} robot spills water while attempting to pour half a cup.
    \item \textit{Slipping objects -} items slip from the robot’s gripper while transferring them between bins.
    \item \textit{Incorrect mixing -} robot mixes rice and beans unevenly.
    \item \textit{Misplaced pouring -} tomato ketchup is poured in the wrong location.
    \item \textit{Collateral events -} robot accidentally knocks over a coffee cup during manipulation.
\end{itemize}

We manually annotated each video with the actual cause of the failure and the corresponding recovery plan, establishing a ground truth for evaluating the model’s predictive accuracy.

To assess failure detection, we compared our model with BLIP2, a state-of-the-art image captioning system. However, BLIP2 struggles with high-level semantic descriptions and often overlooks crucial details required for identifying specific failure modes. Additionally, we analyzed the effect of using sequence-based progressive detection (evaluating multiple frames) versus non-progressive detection (considering only the final frame). As shown in Table \ref{tab:detect_perf}, our progressive detection approach achieves significantly higher performance compared to other methods.

\begin{table}[ht!]
\centering
\caption{Failure detection and explanation accuracy on the dataset}
\label{tab:detect_perf}
\begin{tabular}{|l|c|c|}
\hline
\textbf{Method} & \textbf{Detection (\%)} & \textbf{Explanation (\%)} \\ \hline
BLIP2 Caption & 8 & 2 \\ \hline
Final Frame Only & 82 & 52 \\ \hline
Reflect & 88 & 68 \\ \hline
\textbf{STAR}  & \textbf{90}  & \textbf{78}  \\ \hline
\end{tabular}
\end{table}


In the recovery planning phase, we conducted an ablation study to compare three approaches:

\begin{itemize}
    \item Using only a Knowledge Graph (KG).
    \item Using only a Foundation Model (GPT-4).
    \item STAR, which integrates both the KG and the Foundation Model.
\end{itemize}

The results, shown in Table \ref{tab:recovery_performance}, highlight the advantages of combining structured knowledge with the adaptability of foundation models. Nonetheless, we observed common errors in recovery plans, often generating non-executable tasks, such as instructing the robot to ``collect scattered rice from the table" a task beyond its current capabilities.

\begin{table}[ht!]
\centering
\caption{Recovery plan accuracy}
\label{tab:recovery_performance}
\begin{tabular}{|l|c|}
\hline
\textbf{Method}     & \textbf{Accuracy (\%)} \\ \hline
KG search           &     28    \\ \hline
GPT-4 prompt engineering          &     74    \\ \hline
\textbf{STAR}           & \textbf{78}        \\ \hline
\end{tabular}
\end{table}

\section{Discussion}

In this section, we discuss key insights from our experiments and reflect on the strengths, limitations, and optimization strategies employed during task planning and failure detection. We also highlight potential areas for future improvements and further exploration.

\subsection{Handling Knowledge Graph Initialization}

One of the challenges in robotic task planning, particularly when relying on domain-specific KGs like FOON, is the need to have pre-constructed graphs for every task domain. 
Our framework addresses this limitation by enabling the KG to start from an initially empty state, dynamically growing and updating as the robot interacts with its environment. This adaptive approach ensures that the robot can learn new tasks over time, improving its versatility. Additionally, the framework is designed to integrate with FailNet—a specialized knowledge network focused on failure scenarios—allowing it to accumulate task-specific corrections during operation.

Another strength of our approach is that it supports seamless merging of multiple KGs from different domains. For example, two distinct FOONs or FailNets—one designed for food preparation tasks and another for cleaning tasks—can be easily combined into a unified structure. This flexibility opens the door to multi-domain task planning, where the robot leverages knowledge from various contexts, ultimately enhancing its problem-solving abilities.

\subsection{Optimizing Frame Sampling with Image Grids}

\subsubsection{Challenges of Frame-Based Failure Detection}
Our system relies on multiple video frames to accurately detect and identify failures during task execution. While analyzing multiple frames improves detection accuracy, it introduces two major drawbacks: (1) slower response times from the API due to higher processing requirements, and (2) increased token usage, which raises the cost when interacting with models such as GPT-4.

To address these challenges, we implemented an optimization strategy by creating image grids, which allow multiple frames to be combined into a single image. Instead of sending individual frames to the API, we generate a compact 3x3 grid containing nine frames. This approach significantly reduces token consumption without sacrificing performance. From our experiments, we observed that both methods—using nine individual frames versus a single 3x3 grid—achieved comparable accuracy in failure detection. However, the grid-based approach greatly improved API response time and reduced operational costs, making it a more practical solution for real-time applications.

\subsubsection{Image Grid Example}
\begin{figure}[t!]

	\centering
	\includegraphics[width=\columnwidth]{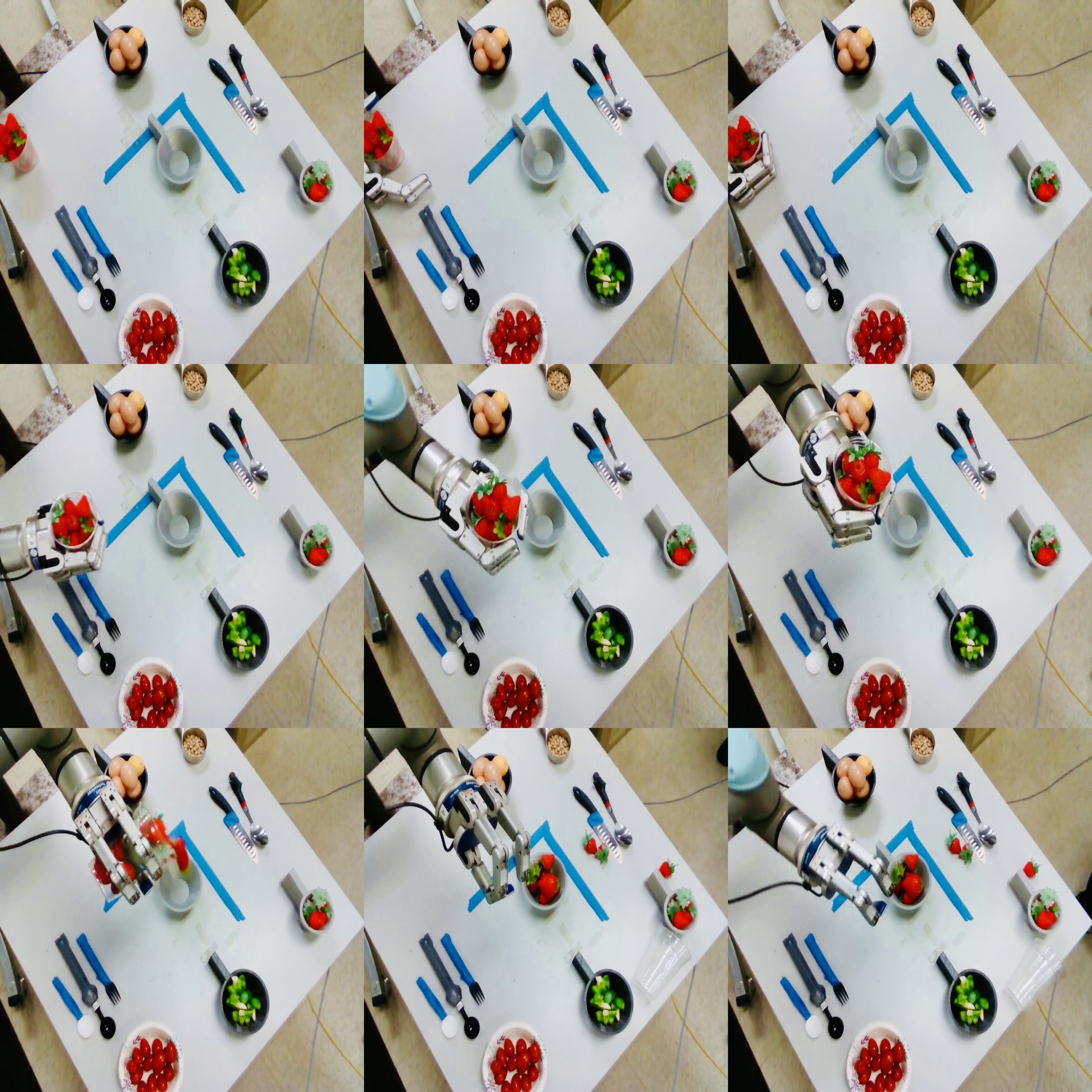}
	\caption{Image grid created with 9 frames to reduce token usage in GPT-4}
	\label{grid}
\end{figure}

Figure \ref{grid} provides an example of the image grid used in our experiments. Each frame in the grid corresponds to a specific moment during task execution, capturing crucial points that help in identifying failures. By consolidating multiple frames into a single input, we minimize API overhead while retaining enough visual information for accurate detection. This optimization paves the way for more efficient failure recovery systems, especially in scenarios that demand rapid response times.

\section{Conclusion}
\label{sec:con}


As robots increasingly integrate into dynamic and unpredictable environments, the need for flexible and resilient systems has become critical. Traditional robots, bound by predefined rules, struggle to adapt when confronted with unforeseen situations. This research addresses these challenges by introducing a hybrid framework STAR (Smart Task Adaptation and Recovery) that combines the generalization capabilities of Foundation Models (FMs) with the structured knowledge storage and reusability of Knowledge Graphs (KGs). STAR bridges the gap between high-level planning and reliable execution, leveraging FMs for tackling novel scenarios and KGs for efficiently storing and organizing task knowledge for future reuse. This dual-layered approach, paired with real-time monitoring and adaptive recovery mechanisms, significantly enhances robots' ability to handle complex tasks and recover from failures autonomously. Experimental results validate the system's effectiveness, demonstrating improvements in adaptability, robustness, and task completion.

Future research will focus on enabling robots to integrate multi-modal inputs—combining vision, sound, and touch (e.g., tactile feedback for spill detection) for richer situational awareness. Additionally, advancing self-supervised learning methods will allow robots to continuously refine their performance through experience, minimizing reliance on extensive labeled datasets. Finally, extending STAR to handle multi-agent scenarios and developing methods for distributed knowledge sharing could enhance its applicability in industrial settings.

\bibliographystyle{IEEEtran}
\bibliography{IEEEabrv,ref}

\end{document}